# Lipschitz Parametrization of Probabilistic Graphical Models

**Jean Honorio**
Computer Science Department
Stony Brook University
Stony Brook, NY 11794

## Abstract

We show that the log-likelihood of several probabilistic graphical models is Lipschitz continuous with respect to the $\ell_p$-norm of the parameters. We discuss several implications of Lipschitz parametrization. We present an upper bound of the Kullback-Leibler divergence that allows understanding methods that penalize the $\ell_p$-norm of differences of parameters as the minimization of that upper bound. The expected log-likelihood is lower bounded by the negative $\ell_p$-norm, which allows understanding the generalization ability of probabilistic models. The exponential of the negative $\ell_p$-norm is involved in the lower bound of the Bayes error rate, which shows that it is reasonable to use parameters as features in algorithms that rely on metric spaces (e.g. classification, dimensionality reduction, clustering). Our results do not rely on specific algorithms for learning the structure or parameters. We show preliminary results for activity recognition and temporal segmentation.

## 1 Introduction

Probabilistic graphical models provide a way to represent variables along with their conditional dependencies and therefore allow formalizing our knowledge of the interacting entities in the real world.

Several methods have been proposed for learning the structure and parameters of graphical models from data. We mention only a few references that follow a maximum likelihood approach for Markov random fields (Lee et al., 2006), Ising models (Höfling & Tibshirani, 2009), Gaussian graphical models (Banerjee et al., 2006; Friedman et al., 2007) and Bayesian networks (Guo & Schuurmans, 2006; Schmidt et al.,

2007). One may ask whether the log-likelihood is "well behaved", i.e. small changes in the parameters produce small changes in the objective function. Another natural question is whether the $\ell_p$ distance between the learnt parameters and the ground truth provides some guarantee on their generalization ability, i.e. the expected log-likelihood.

When learning multiple graphical models, several authors have proposed $\ell_p$-norm regularizers from the difference of parameters between two models. (Zhang & Wang, 2010) proposed a method that detects sparse structural changes of Gaussian graphical models in controlled experiments between two experimental conditions. (Kolar et al., 2010) proposed a total variation regularizer for learning time-varying Ising models with sparse changes along the time course. (Kolar et al., 2009) proposed a similar method for Gaussian graphical models. One natural question is whether the $\ell_p$-norm of the difference of parameters between two graphical models is related to a measure of similarity between probability distributions, i.e. the Kullback-Leibler divergence.

There are several experimental results where the parameters of graphical models were used as features for classification and clustering. Classification of image textures from the precision matrix of Gaussian graphical models as features was proposed in (Chellappa & Chatterjee, 1985), and from parameters of Ising models in (Chen & Dubes, 1990). The use of the covariance matrix as features for detection of humans in still images was proposed in (Tuzel et al., 2007). Clustering by using the Gaussian graphical model parameters was performed in (Kolar et al., 2009), where they show discriminability between different type of imaginations from electroencephalography (EEG) recordings. One may ask whether the parameters of graphical models approximately lie in an metric space ($\ell_p$) that allows for classification and clustering. In other words, whether the $\ell_p$-norm of the difference of parameters between two graphical models is related to a measure

Table 1: Notation used in this paper.

| Notation | Description |
|---|---|
| $\|\mathbf{c}\|_1$ | $\ell_1$-norm of $\mathbf{c} \in \mathbb{R}^N$, i.e. $\sum_n |c_n|$ |
| $\|\mathbf{c}\|_\infty$ | $\ell_\infty$-norm of $\mathbf{c} \in \mathbb{R}^N$, i.e. $\max_n |c_n|$ |
| $\|\mathbf{c}\|_2$ | Euclidean norm of $\mathbf{c} \in \mathbb{R}^N$, i.e. $\sqrt{\sum_n c_n^2}$ |
| $\mathbf{A} \succeq \mathbf{0}$ | $\mathbf{A} \in \mathbb{R}^{N \times N}$ is symmetric and positive semidefinite |
| $\mathbf{A} \succ \mathbf{0}$ | $\mathbf{A} \in \mathbb{R}^{N \times N}$ is symmetric and positive definite |
| $\|\mathbf{A}\|_1$ | $\ell_1$-norm of $\mathbf{A} \in \mathbb{R}^{M \times N}$, i.e. $\sum_{mn} |a_{mn}|$ |
| $\|\mathbf{A}\|_\infty$ | $\ell_\infty$-norm of $\mathbf{A} \in \mathbb{R}^{M \times N}$, i.e. $\max_{mn} |a_{mn}|$ |
| $\|\mathbf{A}\|_2$ | spectral norm of $\mathbf{A} \in \mathbb{R}^{N \times N}$, i.e. the maximum eigenvalue of $\mathbf{A} \succ \mathbf{0}$ |
| $\|\mathbf{A}\|_{\mathfrak{F}}$ | Frobenius norm of $\mathbf{A} \in \mathbb{R}^{M \times N}$, i.e. $\sqrt{\sum_{mn} a_{mn}^2}$ |
| $\langle \mathbf{A}, \mathbf{B} \rangle$ | scalar product of $\mathbf{A}, \mathbf{B} \in \mathbb{R}^{M \times N}$, i.e. $\sum_{mn} a_{mn} b_{mn}$ |
| $\partial f / \partial \mathbf{c}$ | gradient of $f$ with respect to $\mathbf{c} \in \mathbb{R}^N$, i.e. $\partial f / \partial \mathbf{c} \in \mathbb{R}^N$ |
| $\partial f / \partial \mathbf{A}$ | gradient of $f$ with respect to $\mathbf{A} \in \mathbb{R}^{M \times N}$, i.e. $\partial f / \partial \mathbf{A} \in \mathbb{R}^{M \times N}$ |

of discriminability, i.e. the Bayes error rate.

In this paper, we define Lipschitz continuous parametrization of probabilistic models. Through Lipschitz parametrization, we provide an upper bound of the Kullback-Leibler divergence. Therefore, methods that penalize the $\ell_p$-norm of differences of parameters (Kolar et al., 2009; Kolar et al., 2010; Zhang & Wang, 2010) are minimizing an upper bound of the Kullback-Leibler divergence. We show that Lipschitz parametrization also allows understanding the generalization ability of probabilistic models by providing a lower bound of the expected log-likelihood. Finally, we provide a lower bound of the Bayes error rate that depends on the $\ell_p$-norm of the model parameters. This allows understanding the use of model parameters as features for classification and clustering as in (Chellappa & Chatterjee, 1985; Chen & Dubes, 1990; Tuzel et al., 2007; Kolar et al., 2009).

## 2  Preliminaries

In this section, we introduce probabilistic graphical models and Lipschitz continuity. We use the notation in Table 1.

We assume $\mathbf{x} \in \mathbb{R}^N$ for continuous random variables. For discrete random variables, we assume $\mathbf{x} \in \times_n \{1, \ldots, X_n\}$, i.e. $(\forall n) \; x_n \in \{1, \ldots, X_n\}$. First, we define three general classes of graphical models: *Bayesian networks*, *Markov random fields* and *factor graphs*.

**Definition 1.** *A Bayesian network (Koller & Friedman, 2009; Lauritzen, 1996) for random variables $\mathbf{x}$ is a directed acyclic graph with one conditional proba-* bility function $p(x_n | \mathbf{x}_{\pi_n})$ for each variable $x_n$ given its set of parents $\pi_n \subseteq \{1, \ldots, N\}$. The joint probability distribution is given by:

$$p(\mathbf{x}) = \prod_n p(x_n | \mathbf{x}_{\pi_n}) \quad (1)$$

where $(\forall n, \mathbf{x}_{\pi_n}) \int_{x_n} p(x_n | \mathbf{x}_{\pi_n}) = 1$ and therefore $p(\mathbf{x})$ is valid, i.e. $\int_\mathbf{x} p(\mathbf{x}) = 1$.

**Definition 2.** *A Markov random field (Koller & Friedman, 2009; Lauritzen, 1996) for random variables $\mathbf{x}$ is an undirected graph with one potential function $\phi_c$ for each maximal clique $\varphi_c \subseteq \{1, \ldots, N\}$. The joint probability distribution is given by:*

$$p(\mathbf{x}) = \frac{1}{\mathcal{Z}} \prod_c \phi_c(\mathbf{x}_{\varphi_c}) \quad (2)$$

where the partition function $\mathcal{Z} = \int_\mathbf{x} \prod_c \phi_c(\mathbf{x}_{\varphi_c})$ ensures that $p(\mathbf{x})$ is valid, i.e. $\int_\mathbf{x} p(\mathbf{x}) = 1$.

**Definition 3.** *A factor graph (Koller & Friedman, 2009) for random variables $\mathbf{x}$ is a bipartite graph where one set of nodes are the random variables and the other set are the local functions. Each local function $\phi_c$ is connected to the set variables of variables $\varphi_c \subseteq \{1, \ldots, N\}$ on which it depends on. The joint probability distribution is given by:*

$$p(\mathbf{x}) = \frac{1}{\mathcal{Z}} \prod_c \phi_c(\mathbf{x}_{\varphi_c}) \quad (3)$$

where the partition function $\mathcal{Z} = \int_\mathbf{x} \prod_c \phi_c(\mathbf{x}_{\varphi_c})$ ensures that $p(\mathbf{x})$ is valid, i.e. $\int_\mathbf{x} p(\mathbf{x}) = 1$.

For completeness, we introduce Lipschitz continuity for differentiable functions.

**Definition 4.** *Given the parameters $\boldsymbol{\Theta} \in \mathbb{R}^{M_1 \times M_2}$, a differentiable function $f(\boldsymbol{\Theta}) \in \mathbb{R}$ is called Lipschitz continuous with respect to the $\ell_p$-norm of $\boldsymbol{\Theta}$, if there exists a constant $K \geq 0$ such that:*

$$(\forall \boldsymbol{\Theta}_1, \boldsymbol{\Theta}_2) \; |f(\boldsymbol{\Theta}_1) - f(\boldsymbol{\Theta}_2)| \leq K \|\boldsymbol{\Theta}_1 - \boldsymbol{\Theta}_2\|_p \quad (4)$$

or equivalently:

$$(\forall \boldsymbol{\Theta}) \; \|\partial f / \partial \boldsymbol{\Theta}\|_p \leq K \quad (5)$$

## 3  Lipschitz Parametrization and Implications

In this section, we define Lipschitz parametrization of probabilistic models and discuss its implications.

### 3.1  Lipschitz Parametrization

We extend the Lipschitz continuity notion to the parametrization of probability distributions.

**Definition 5.** *A probability distribution $\mathcal{P} = p(\cdot|\boldsymbol{\Theta})$ parameterized by $\boldsymbol{\Theta} \in \mathbb{R}^{M_1 \times M_2}$ is called $(\ell_p, K)$-Lipschitz continuous if for all $\mathbf{x}$, the log-likelihood $f(\boldsymbol{\Theta}) = \log p(\mathbf{x}|\boldsymbol{\Theta})$ is Lipschitz continuous with respect to the $\ell_p$-norm of $\boldsymbol{\Theta}$ with constant $K(\mathbf{x})$.*

**Remark 6.** *Note that $(\ell_p, K)$-Lipschitz continuity implies $(\ell_{p'}, K')$-Lipschitz continuity, since all vector and matrix norms are equivalent, i.e. $(\forall \boldsymbol{\Theta} \in \mathbb{R}^{M_1 \times M_2})$ $\alpha \|\boldsymbol{\Theta}\|_p \leq \|\boldsymbol{\Theta}\|_{p'} \leq \beta \|\boldsymbol{\Theta}\|_p$ for some $\alpha, \beta > 0$ and $M_1, M_2 < +\infty$.*

If we are interested in Euclidean spaces, we would need to prove Lipschitz continuity with respect to the $\ell_2$-norm for vectors or the Frobenius norm for matrices. Due to Remark 6, we can chose any particular norm for proving Lipschitz continuity.

### 3.2 Kullback-Leibler Divergence

We show that the $\ell_p$-norm is an upper bound of the Kullback-Leibler divergence.

**Theorem 7.** *Given two $(\ell_p, K)$-Lipschitz continuous distributions $\mathcal{P}_1 = p(\cdot|\boldsymbol{\Theta}_1)$ and $\mathcal{P}_2 = p(\cdot|\boldsymbol{\Theta}_2)$, the Kullback-Leibler divergence from $\mathcal{P}_1$ to $\mathcal{P}_2$ is bounded as follows:*

$$\mathcal{KL}(\mathcal{P}_1 \| \mathcal{P}_2) \leq \overline{K} \|\boldsymbol{\Theta}_1 - \boldsymbol{\Theta}_2\|_p \quad (6)$$

*with constant $\overline{K} = \mathbb{E}_{\mathcal{P}_1}[K(\mathbf{x})]$.*

*Proof.* By definition $\mathcal{KL}(\mathcal{P}_1 \| \mathcal{P}_2) = \mathbb{E}_{\mathcal{P}_1}[\log p(\mathbf{x}|\boldsymbol{\Theta}_1) - \log p(\mathbf{x}|\boldsymbol{\Theta}_2)] \leq \mathbb{E}_{\mathcal{P}_1}[|\log p(\mathbf{x}|\boldsymbol{\Theta}_1) - \log p(\mathbf{x}|\boldsymbol{\Theta}_2)|] \equiv B$. Note that by Definitions 4 and 5, $B \leq \mathbb{E}_{\mathcal{P}_1}[K(\mathbf{x})\|\boldsymbol{\Theta}_1 - \boldsymbol{\Theta}_2\|_p] = \mathbb{E}_{\mathcal{P}_1}[K(\mathbf{x})]\|\boldsymbol{\Theta}_1 - \boldsymbol{\Theta}_2\|_p = \overline{K}\|\boldsymbol{\Theta}_1 - \boldsymbol{\Theta}_2\|_p$. ☐

**Remark 8.** *For identifiable distributions $\mathcal{P}_1 = p(\cdot|\boldsymbol{\Theta}_1)$ and $\mathcal{P}_2 = p(\cdot|\boldsymbol{\Theta}_2)$ (i.e. $\mathcal{P}_1 = \mathcal{P}_2 \Rightarrow \boldsymbol{\Theta}_1 = \boldsymbol{\Theta}_2$), the upper bound in Theorem 7 is tight since the Kullback-Leibler divergence is zero if and only if the parameters are equal. More formally, $\mathcal{KL}(\mathcal{P}_1 \| \mathcal{P}_2) = 0 \Leftrightarrow \mathcal{P}_1 = \mathcal{P}_2 \Leftrightarrow \boldsymbol{\Theta}_1 = \boldsymbol{\Theta}_2 \Leftrightarrow \|\boldsymbol{\Theta}_1 - \boldsymbol{\Theta}_2\|_p = 0$.*

**Remark 9.** *The upper bound in Theorem 7 also applies for every marginal distribution by properties of the Kullback-Leibler divergence.*

### 3.3 Expected Log-Likelihood

We show that the negative $\ell_p$-norm is a lower bound of the expected log-likelihood.

**Theorem 10.** *Given two $(\ell_p, K)$-Lipschitz continuous distributions $\mathcal{P} = p(\cdot, \boldsymbol{\Theta})$ and $\mathcal{P}^* = p(\cdot, \boldsymbol{\Theta}^*)$, the expected log-likelihood (also called negative cross entropy) of the learnt distribution $\mathcal{P}$ with respect to the ground truth distribution $\mathcal{P}^*$ is bounded as follows:*

$$-\mathcal{H}(\mathcal{P}^*) - \overline{K}\|\boldsymbol{\Theta}^* - \boldsymbol{\Theta}\|_p \leq \mathbb{E}_{\mathcal{P}^*}[\log p(\mathbf{x}|\boldsymbol{\Theta})] \leq 0 \quad (7)$$

*with constant $\overline{K} = \mathbb{E}_{\mathcal{P}^*}[K(\mathbf{x})]$.*

*Proof.* Note that $0 = -\mathcal{H}(\mathcal{P}^*) - \mathbb{E}_{\mathcal{P}^*}[\log p(\mathbf{x}|\boldsymbol{\Theta}^*)]$ and therefore $\mathbb{E}_{\mathcal{P}^*}[\log p(\mathbf{x}|\boldsymbol{\Theta})] = \mathbb{E}_{\mathcal{P}^*}[\log p(\mathbf{x}|\boldsymbol{\Theta})] - \mathcal{H}(\mathcal{P}^*) - \mathbb{E}_{\mathcal{P}^*}[\log p(\mathbf{x}|\boldsymbol{\Theta}^*)] = -\mathcal{H}(\mathcal{P}^*) - \mathbb{E}_{\mathcal{P}^*}[\log p(\mathbf{x}|\boldsymbol{\Theta}^*) - \log p(\mathbf{x}|\boldsymbol{\Theta})] = -\mathcal{H}(\mathcal{P}^*) - \mathcal{KL}(\mathcal{P}^* \| \mathcal{P})$. The upper bound follows from the non-negativity of the Kullback-Leibler divergence and entropy.

For proving the lower bound, given that $\mathcal{KL}(\mathcal{P}^* \| \mathcal{P}) \leq \overline{K}\|\boldsymbol{\Theta}^* - \boldsymbol{\Theta}\|_p$ by Theorem 7, we prove our claim. ☐

In the following Section 4, we prove that for probabilistic models over discrete random variables, $(\forall \mathbf{x})$ $K(\mathbf{x}) = 1$ and therefore $\overline{K} = 1$. For continuous random variables, given its generality, the constant $K(\mathbf{x})$ depends on $\mathbf{x}$ and therefore $\overline{K}$ is looser and does not have a closed-form expression; except for specific cases, e.g. Gaussian graphical models.

### 3.4 Bayes Error Rate

We show that that the exponential of the negative $\ell_p$-norm is involved in the lower bound of the Bayes error rate. We also motivate a distance measure similar to the Chernoff bound (Chernoff, 1952), i.e. the negative log-Bayes error rate.

**Theorem 11.** *Given two classes $\varpi_1$ and $\varpi_2$ with priors $P(\varpi_1) = P(\varpi_2) = \frac{1}{2}$ and their corresponding $(\ell_p, K)$-Lipschitz continuous distributions $\mathcal{P}_1 = p(\cdot|\boldsymbol{\Theta}_1)$ and $\mathcal{P}_2 = p(\cdot|\boldsymbol{\Theta}_2)$, the Bayes error rate $\mathcal{BE}(\boldsymbol{\Theta}_1, \boldsymbol{\Theta}_2) = \frac{1}{2}\int_{\mathbf{x}} \min(p(\mathbf{x}|\boldsymbol{\Theta}_1), p(\mathbf{x}|\boldsymbol{\Theta}_2))$ is bounded as follows:*

$$\frac{\mathcal{BB}(\boldsymbol{\Theta}_1, \boldsymbol{\Theta}_2)}{4} \leq \mathcal{BE}(\boldsymbol{\Theta}_1, \boldsymbol{\Theta}_2) \quad (8)$$

$$\log 2 \leq -\log \mathcal{BE}(\boldsymbol{\Theta}_1, \boldsymbol{\Theta}_2) \leq \log 4 + \widetilde{K}\|\boldsymbol{\Theta}_1 - \boldsymbol{\Theta}_2\|_p \quad (9)$$

*where $\mathcal{BB}(\boldsymbol{\Theta}_1, \boldsymbol{\Theta}_2) = \sum_c \mathbb{E}_{\mathcal{P}_c}[e^{-K(\mathbf{x})\|\boldsymbol{\Theta}_1 - \boldsymbol{\Theta}_2\|_p}]$ and $\widetilde{K} = \min_c \mathbb{E}_{\mathcal{P}_c}[K(\mathbf{x})]$.*

*Proof.* Let $p_c \equiv p(\mathbf{x}|\boldsymbol{\Theta}_c)$. We can rewrite $\mathcal{BE}(\boldsymbol{\Theta}_1, \boldsymbol{\Theta}_2) = \frac{1}{2}\int_{\mathbf{x}} \min\left(\frac{p_1}{p_1 + p_2}, \frac{p_2}{p_1 + p_2}\right)(p_1 + p_2) = \frac{1}{2}\int_{\mathbf{x}} e^{\min\left(\log\frac{p_1}{p_1 + p_2}, \log\frac{p_2}{p_1 + p_2}\right)}(p_1 + p_2)$. We can also rewrite $\log \frac{p_1}{p_1 + p_2} = -\log\left(1 + \frac{p_2}{p_1}\right) = -\ell(z_{12})$, where $z_{12} = \log p_1 - \log p_2$ and $\ell(z) = \log(1 + e^{-z})$ is the logistic loss. Similarly $\log \frac{p_2}{p_1 + p_2} = -\ell(-z_{12})$. Therefore $\min\left(\log\frac{p_1}{p_1 + p_2}, \log\frac{p_2}{p_1 + p_2}\right) = \min(-\ell(z_{12}), -\ell(-z_{12}))$. Note that $(\forall z) -|z| - \log 2 \leq \min(-\ell(z), -\ell(-z))$. Since both $\mathcal{P}_1$ and $\mathcal{P}_2$ are $(\ell_p, K)$-Lipschitz continuous, by Definitions 4 and 5, we have $-K(\mathbf{x})\|\boldsymbol{\Theta}_1 - \boldsymbol{\Theta}_2\|_p - \log 2 \leq \min\left(\log\frac{p_1}{p_1 + p_2}, \log\frac{p_2}{p_1 + p_2}\right)$.

For proving the lower bound in eq.(8), $\mathcal{BE}(\boldsymbol{\Theta}_1, \boldsymbol{\Theta}_2) \geq \frac{1}{2}\int_{\mathbf{x}} e^{-K(\mathbf{x})\|\boldsymbol{\Theta}_1 - \boldsymbol{\Theta}_2\|_p - \log 2}(p_1 + p_2) = \frac{1}{4}\int_{\mathbf{x}} e^{-K(\mathbf{x})\|\boldsymbol{\Theta}_1 - \boldsymbol{\Theta}_2\|_p}(p_1 + p_2) = \frac{1}{4}\mathcal{BB}(\boldsymbol{\Theta}_1, \boldsymbol{\Theta}_2)$.

The lower bound of eq.(9) follows from the fact that $\mathcal{BE}(\Theta_1, \Theta_2) \leq \frac{1}{2}$. For proving the upper bound of eq.(9), by Jensen's inequality $\frac{1}{4}\sum_c e^{-\mathbb{E}_{\mathcal{P}_c}[K(\mathbf{x})]\|\Theta_1 - \Theta_2\|_p} \leq \frac{\mathcal{BB}(\Theta_1, \Theta_2)}{4} \leq \mathcal{BE}(\Theta_1, \Theta_2)$. Therefore, $-\log \mathcal{BE}(\Theta_1, \Theta_2) \leq \log 4 - \log \sum_c e^{-\mathbb{E}_{\mathcal{P}_c}[K(\mathbf{x})]\|\Theta_1 - \Theta_2\|_p}$. By properties of the logsumexp function, we have $-\log \mathcal{BE}(\Theta_1, \Theta_2) \leq \log 4 - \max_c (-\mathbb{E}_{\mathcal{P}_c}[K(\mathbf{x})])\|\Theta_1 - \Theta_2\|_p = \log 4 + \min_c \mathbb{E}_{\mathcal{P}_c}[K(\mathbf{x})]\|\Theta_1 - \Theta_2\|_p$. □

# 4 Lipschitz Continuous Models

In this section, we show that several probabilistic graphical models are Lipschitz continuous. This includes Bayesian networks, Markov random fields and factor graphs for discrete and continuous random variables. Dynamic models such as dynamic Bayesian networks and conditional random fields are also Lipschitz continuous.

## 4.1 Bayesian Networks

We show that a sufficient condition for the Lipschitz continuity of Bayesian networks is the Lipschitz continuity of the conditional probability functions.

**Lemma 12.** *Given a $(\ell_p, K)$-Lipschitz continuous conditional probability function $p(x_n|\mathbf{x}_{\pi_n}, \Theta)$ for each variable $x_n$, the Bayesian network $p(\mathbf{x}|\Theta) = \prod_n p(x_n|\mathbf{x}_{\pi_n}, \Theta)$ is $(\ell_p, NK)$-Lipschitz continuous.*

*Proof.* Let $g_n(\Theta) = \log p(x_n|\mathbf{x}_{\pi_n}, \Theta)$ and $f(\Theta) = \log p(\mathbf{x}|\Theta) = \sum_n \log p(x_n|\mathbf{x}_{\pi_n}, \Theta) = \sum_n g_n(\Theta)$, and therefore $\partial f/\partial\Theta = \sum_n \partial g_n/\partial\Theta$. By Definitions 4 and 5, we have $\|\partial f/\partial\Theta\|_p \leq \sum_n \|\partial g_n/\partial\Theta\|_p \leq NK(\mathbf{x})$. □

**Remark 13.** *When comparing two Bayesian networks, the set of parents $\pi_n$ for each variable $x_n$ is not necessarily the same for both networks. Since Lemma 12 does not use the fact that the joint probability distribution $p(\mathbf{x}|\Theta)$ is valid (i.e. $\int_{\mathbf{x}} p(\mathbf{x}|\Theta) = 1$ which is given by the acyclicity constraints), we can join the set of parents of both Bayesian networks before comparing them. More formally, let $\pi_n^{(1)}$ and $\pi_n^{(2)}$ be the set of parents of variable $x_n$ in Bayesian network 1 and 2 respectively. It is trivial to show that if $p(x_n|\mathbf{x}_{\pi_n^{(1)}}, \Theta)$ and $p(x_n|\mathbf{x}_{\pi_n^{(2)}}, \Theta)$ are Lipschitz continuous, so is $p(x_n|\mathbf{x}_{\pi_n^{(1)} \cup \pi_n^{(2)}}, \Theta)$.*

Given the previous discussion, in the sequel, we show Lipschitz continuity for the conditional probability functions only.

## 4.2 Discrete Bayesian Networks

The following parametrization of Bayesian networks for discrete random variables is equivalent to using conditional probability tables. We use a representation in an exponential space that resembles the *softmax activation function* in the neural networks literature (Duda et al., 2001).

**Lemma 14.** *Let $x_{\pi_n}$ be one of the possible parent value combinations for variable $x_n$, i.e. $x_{\pi_n} \in \{1, \ldots, X_{\pi_n}\}$ where $X_{\pi_n} = \prod_{n' \in \pi_n} X_{n'}$. The conditional probability mass function for the discrete Bayesian network parameterized by $\Theta = \{\mathbf{w}^{(n,1)}, \ldots, \mathbf{w}^{(n,X_{\pi_n})}\}_n$, $(\forall n, x_{\pi_n})\ \mathbf{w}^{(n,x_{\pi_n})} \in \mathbb{R}^{X_n - 1}$:*

$$\mathbb{P}[x_n = i | x_{\pi_n} = j, \Theta] = \frac{e^{w_i^{(n,j)}1[i < X_n]}}{\sum_{x_n} e^{w_{x_n}^{(n,j)}} + 1} \qquad (10)$$

*is $(\ell_\infty, 1)$-Lipschitz continuous.*

*Proof.* Let $\mathbf{w} \equiv \mathbf{w}^{(n,x_{\pi_n})}$. For $i < X_n$, let $f(\mathbf{w}) = \log \mathbb{P}[x_n = i | x_{\pi_n} = j, \Theta] = w_i - \log(\sum_{x_n} e^{w_{x_n}} + 1)$. By deriving $\partial f/\partial w_i = 1 - \frac{e^{w_i}}{\sum_{x_n} e^{w_{x_n}} + 1} = 1 - \mathbb{P}[x_n = i | x_{\pi_n} = j, \Theta]$. Since $(\forall i)\ 0 \leq \mathbb{P}[x_n = i | x_{\pi_n} = j, \Theta] \leq 1$, it follows that $|\partial f/\partial w_i| \leq 1$ and therefore $\|\partial f/\partial\mathbf{w}\|_\infty \leq 1$. By Definitions 4 and 5 we prove our claim. □

The following parametrization of Bayesian networks for discrete random variables corresponds to the *multinomial logistic regression*. It reduces to logistic regression for binary variables.

**Lemma 15.** *Given a feature function with $F$ features $\psi(\mathbf{x}_{\pi_n}) = (\psi_1(\mathbf{x}_{\pi_n}), \ldots, \psi_F(\mathbf{x}_{\pi_n}))^{\mathrm{T}}$ such that $(\forall \mathbf{x}_{\pi_n})\ \|\psi(\mathbf{x}_{\pi_n})\|_\infty \leq 1$, the conditional probability mass function for the discrete Bayesian network parameterized by $\Theta = \{\mathbf{w}_{(1)}^{(n)}, \ldots, \mathbf{w}_{(X_n - 1)}^{(n)}\}_n$, $(\forall n, x_n)\ \mathbf{w}_{(x_n)}^{(n)} \in \mathbb{R}^F$:*

$$\mathbb{P}[x_n = i | \mathbf{x}_{\pi_n}, \Theta] = \frac{e^{\mathbf{w}_{(i)}^{(n)\mathrm{T}}\psi(\mathbf{x}_{\pi_n})1[i < X_n]}}{\sum_{x_n} e^{\mathbf{w}_{(x_n)}^{(n)\mathrm{T}}\psi(\mathbf{x}_{\pi_n})} + 1} \qquad (11)$$

*is $(\ell_\infty, 1)$-Lipschitz continuous.*

*Proof.* Let $\mathbf{w} \equiv \mathbf{w}^{(n)}$. For $i < X_n$, let $f(\mathbf{w}) = \mathbb{P}[x_n = i | \mathbf{x}_{\pi_n}, \Theta] = \mathbf{w}_{(i)}^{\mathrm{T}}\psi(\mathbf{x}_{\pi_n}) - \log(\sum_{x_n} e^{\mathbf{w}_{(x_n)}^{\mathrm{T}}\psi(\mathbf{x}_{\pi_n})} + 1)$. By deriving $\partial f/\partial\mathbf{w}_{(i)} = \psi(\mathbf{x}_{\pi_n}) - \frac{e^{\mathbf{w}_{(i)}^{\mathrm{T}}\psi(\mathbf{x}_{\pi_n})}\psi(\mathbf{x}_{\pi_n})}{\sum_{x_n} e^{\mathbf{w}_{(x_n)}^{\mathrm{T}}\psi(\mathbf{x}_{\pi_n})} + 1} = (1 - \mathbb{P}[x_n = i | \mathbf{x}_{\pi_n}, \Theta])\psi(\mathbf{x}_{\pi_n})$. Since $(\forall i)\ 0 \leq \mathbb{P}[x_n = i | \mathbf{x}_{\pi_n}, \Theta] \leq 1$, it follows that $\|\partial f/\partial\mathbf{w}_{(i)}\|_\infty \leq \|\psi(\mathbf{x}_{\pi_n})\|_\infty \leq 1$. By Definitions 4 and 5, we prove our claim. □

Note that the requirement that $(\forall \mathbf{x})\ \|\boldsymbol{\psi}(\mathbf{x}_{\pi_n})\|_\infty \leq 1$ is not restrictive, since the random variables are discrete and we can perform scaling of the features.

### 4.3 Continuous Bayesian Networks

We focus on two types of continuous random variables: Gaussian and Laplace. For the Gaussian Bayesian network, we assume that the weight vector $\mathbf{w}$ of linear regression has bounded norm, i.e. $\|\mathbf{w}\|_2 \leq \beta$ (please, see Appendix A[1]). We also assume that the features are normalized, i.e. the standard deviation is one.

**Lemma 16.** *Given a feature function with $F$ features $\boldsymbol{\psi}(\mathbf{x}_{\pi_n}) = (\psi_1(\mathbf{x}_{\pi_n}), \ldots, \psi_F(\mathbf{x}_{\pi_n}))^{\mathrm{T}}$, the conditional probability density function for the Gaussian Bayesian network parameterized by $\boldsymbol{\Theta} = \{\mathbf{w}^{(n)}\}_n$, $(\forall n)\ \mathbf{w}^{(n)} \in \mathbb{R}^F$:*

$$p(x_n|\mathbf{x}_{\pi_n}, \boldsymbol{\Theta}) = \frac{1}{\sqrt{2\pi}} e^{-\frac{1}{2}(x_n - \mathbf{w}^{(n)\mathrm{T}}\boldsymbol{\psi}(\mathbf{x}_{\pi_n}))^2} \quad (12)$$

*is $(\ell_2, \|\boldsymbol{\psi}(\mathbf{x}_{\pi_n})\|_2|x_n| + \beta\|\boldsymbol{\psi}(\mathbf{x}_{\pi_n})\|_2^2)$-Lipschitz continuous.*

*Proof.* Let $\mathbf{w} \equiv \mathbf{w}^{(n)}$ and $f(\mathbf{w}) = \log p(x_n|\mathbf{x}_{\pi_n}, \boldsymbol{\Theta}) = \frac{1}{2}(-\log(2\pi) - (x_n - \mathbf{w}^{\mathrm{T}}\boldsymbol{\psi}(\mathbf{x}_{\pi_n}))^2)$. By deriving $\partial f/\partial \mathbf{w} = (x_n - \mathbf{w}^{\mathrm{T}}\boldsymbol{\psi}(\mathbf{x}_{\pi_n}))\boldsymbol{\psi}(\mathbf{x}_{\pi_n})$. Therefore $\|\partial f/\partial \mathbf{w}\|_2 \leq |x_n - \mathbf{w}^{\mathrm{T}}\boldsymbol{\psi}(\mathbf{x}_{\pi_n})|\ \|\boldsymbol{\psi}(\mathbf{x}_{\pi_n})\|_2 \leq (|x_n| + |\mathbf{w}^{\mathrm{T}}\boldsymbol{\psi}(\mathbf{x}_{\pi_n})|)\ \|\boldsymbol{\psi}(\mathbf{x}_{\pi_n})\|_2 \leq (|x_n| + \|\mathbf{w}\|_2\|\boldsymbol{\psi}(\mathbf{x}_{\pi_n})\|_2)\ \|\boldsymbol{\psi}(\mathbf{x}_{\pi_n})\|_2$. By noting that $\|\mathbf{w}\|_2 \leq \beta$ and by Definitions 4 and 5, we prove our claim. $\square$

**Remark 17.** *In Lemma 16, the expression $K(\mathbf{x}) = \|\boldsymbol{\psi}(\mathbf{x}_{\pi_n})\|_2|x_n| + \beta\|\boldsymbol{\psi}(\mathbf{x}_{\pi_n})\|_2^2$ becomes more familiar for a linear feature function $\boldsymbol{\psi}(\mathbf{x}_{\pi_n}) = \mathbf{x}_{\pi_n}$. In this case, note that $(\forall n)\ \|\boldsymbol{\psi}(\mathbf{x}_{\pi_n})\|_2 = \|\mathbf{x}_{\pi_n}\|_2 \leq \|\mathbf{x}\|_2$ and $(\forall n)\ |x_n| \leq \|\mathbf{x}\|_2$. Therefore $K(\mathbf{x}) \leq (1+\beta)\|\mathbf{x}\|_2^2$.*

For the Laplace Bayesian network, we assume that the features are normalized, i.e. the absolute deviation is one.

**Lemma 18.** *Given a feature function with $F$ features $\boldsymbol{\psi}(\mathbf{x}_{\pi_n}) = (\psi_1(\mathbf{x}_{\pi_n}), \ldots, \psi_F(\mathbf{x}_{\pi_n}))^{\mathrm{T}}$, the conditional probability density function for the Laplace Bayesian network parameterized by $\boldsymbol{\Theta} = \{\mathbf{w}^{(n)}\}_n$, $(\forall n)\ \mathbf{w}^{(n)} \in \mathbb{R}^F$:*

$$p(x_n|\mathbf{x}_{\pi_n}, \boldsymbol{\Theta}) = \frac{1}{2}e^{-|x_n - \mathbf{w}^{(n)\mathrm{T}}\boldsymbol{\psi}(\mathbf{x}_{\pi_n})|} \quad (13)$$

*is $(\ell_2, \|\boldsymbol{\psi}(\mathbf{x}_{\pi_n})\|_2)$-Lipschitz continuous.*

*Proof.* Let $\mathbf{w} \equiv \mathbf{w}^{(n)}$ and $f(\mathbf{w}) = \log p(x_n|\mathbf{x}_{\pi_n}, \boldsymbol{\Theta}) = -\log 2 - |x_n - \mathbf{w}^{\mathrm{T}}\boldsymbol{\psi}(\mathbf{x}_{\pi_n})|$. The subdifferential set of the non-smooth function $f$ can be written as $\partial f/\partial \mathbf{w} =$

$\boldsymbol{\psi}(\mathbf{x}_{\pi_n})s(x_n - \mathbf{w}^{\mathrm{T}}\boldsymbol{\psi}(\mathbf{x}_{\pi_n}))$, where $s(z) = +1$ for $z > 0$, $s(z) = -1$ for $z < 0$ and $s(z) \in [-1; +1]$ for $z = 0$. Therefore $\|\partial f/\partial \mathbf{w}\|_2 \leq \|\boldsymbol{\psi}(\mathbf{x}_{\pi_n})\|_2$. By Definitions 4 and 5, we prove our claim. $\square$

### 4.4 Discrete Factor Graphs

The following parameterization of factor graphs for discrete random variables includes Markov random fields when the features depend on the cliques. A special case of this parametrization are Ising models (i.e. Markov random fields on binary variables with pairwise interactions). The feature function $\boldsymbol{\psi}(\mathbf{x}) = (\mathbf{vec}(\mathbf{x}\mathbf{x}^{\mathrm{T}}), \mathbf{x})$ for Ising models with external field, and $\boldsymbol{\psi}(\mathbf{x}) = \mathbf{vec}(\mathbf{x}\mathbf{x}^{\mathrm{T}})$ without external field.

**Lemma 19.** *Given a feature function with $F$ features $\boldsymbol{\psi}(\mathbf{x}) = (\psi_1(\mathbf{x}), \ldots, \psi_F(\mathbf{x}))^{\mathrm{T}}$ such that $(\forall \mathbf{x})\ \|\boldsymbol{\psi}(\mathbf{x})\|_\infty \leq 1$, the discrete factor graph $\mathcal{P} = p(\cdot, \boldsymbol{\Theta})$ parameterized by $\boldsymbol{\Theta} = \mathbf{w}$, $\mathbf{w} \in \mathbb{R}^F$ with probability mass function:*

$$p(\mathbf{x}|\boldsymbol{\Theta}) = \frac{1}{\mathcal{Z}(\mathbf{w})}e^{\mathbf{w}^{\mathrm{T}}\boldsymbol{\psi}(\mathbf{x})} \quad (14)$$

*where $\mathcal{Z}(\mathbf{w}) = \sum_{\mathbf{x}} e^{\mathbf{w}^{\mathrm{T}}\boldsymbol{\psi}(\mathbf{x})}$ is $(\ell_\infty, 2)$-Lipschitz continuous.*

*Proof.* Let $f(\mathbf{w}) = \log p(\mathbf{x}|\boldsymbol{\Theta}) = \mathbf{w}^{\mathrm{T}}\boldsymbol{\psi}(\mathbf{x}) - \log(\sum_{\mathbf{x}} e^{\mathbf{w}^{\mathrm{T}}\boldsymbol{\psi}(\mathbf{x})})$. By deriving $\partial f/\partial \mathbf{w} = \boldsymbol{\psi}(\mathbf{x}) - \frac{\sum_{\mathbf{x}} e^{\mathbf{w}^{\mathrm{T}}\boldsymbol{\psi}(\mathbf{x})}\boldsymbol{\psi}(\mathbf{x})}{\sum_{\mathbf{x}} e^{\mathbf{w}^{\mathrm{T}}\boldsymbol{\psi}(\mathbf{x})}} = \boldsymbol{\psi}(\mathbf{x}) - \mathbb{E}_{\mathcal{P}}[\boldsymbol{\psi}(\mathbf{x})]$. Since the expected value for discrete random variables is a weighted sum with positive weights that add up to 1 and $(\forall \mathbf{x})\ \|\boldsymbol{\psi}(\mathbf{x})\|_\infty \leq 1$ therefore $\|\mathbb{E}_{\mathcal{P}}[\boldsymbol{\psi}(\mathbf{x})]\|_\infty \leq 1$. It follows that $\|\partial f/\partial \mathbf{w}\|_\infty \leq \|\boldsymbol{\psi}(\mathbf{x})\|_\infty + \|\mathbb{E}_{\mathcal{P}}[\boldsymbol{\psi}(\mathbf{x})]\|_\infty \leq 2$. By Definitions 4 and 5, we prove our claim. $\square$

Note that the requirement that $(\forall \mathbf{x})\ \|\boldsymbol{\psi}(\mathbf{x})\|_\infty \leq 1$ is not restrictive, since the random variables are discrete and we can perform scaling of the features.

### 4.5 Continuous Factor Graphs

The following parameterization of factor graphs for continuous random variables includes Markov random fields when the features depend on the cliques. A special case of this parametrization are Gaussian graphical models (i.e. Markov random fields on jointly Gaussian variables), in which the feature function $\boldsymbol{\psi}(\mathbf{x}) = \mathbf{vec}(\mathbf{x}\mathbf{x}^{\mathrm{T}})$.

**Lemma 20.** *Given a feature function with $F$ features $\boldsymbol{\psi}(\mathbf{x}) = (\psi_1(\mathbf{x}), \ldots, \psi_F(\mathbf{x}))^{\mathrm{T}}$ such that $\mathbb{E}_{\mathcal{P}}[\|\boldsymbol{\psi}(\mathbf{x})\|_p] \leq \alpha$, the continuous factor graph $\mathcal{P} = p(\cdot, \boldsymbol{\Theta})$ parameterized by $\boldsymbol{\Theta} = \mathbf{w}$, $\mathbf{w} \in \mathbb{R}^F$ with probability density function:*

$$p(\mathbf{x}|\boldsymbol{\Theta}) = \frac{1}{\mathcal{Z}(\mathbf{w})}e^{\mathbf{w}^{\mathrm{T}}\boldsymbol{\psi}(\mathbf{x})} \quad (15)$$

---

[1] Appendices are included in the supplementary material at http://www.cs.sunysb.edu/~jhonorio/

where $\mathcal{Z}(\mathbf{w}) = \int_{\mathbf{x}} e^{\mathbf{w}^{\mathrm{T}} \boldsymbol{\psi}(\mathbf{x})}$ is $(\ell_p, \|\boldsymbol{\psi}(\mathbf{x})\|_p + \alpha)$-Lipschitz continuous.

*Proof.* Let $f(\mathbf{w}) = \log p(\mathbf{x}|\boldsymbol{\Theta}) = \mathbf{w}^{\mathrm{T}} \boldsymbol{\psi}(\mathbf{x}) - \log(\int_{\mathbf{x}} e^{\mathbf{w}^{\mathrm{T}} \boldsymbol{\psi}(\mathbf{x})})$. By deriving $\partial f / \partial \mathbf{w} = \boldsymbol{\psi}(\mathbf{x}) - \frac{\int_{\mathbf{x}} e^{\mathbf{w}^{\mathrm{T}} \boldsymbol{\psi}(\mathbf{x})} \boldsymbol{\psi}(\mathbf{x})}{\int_{\mathbf{x}} e^{\mathbf{w}^{\mathrm{T}} \boldsymbol{\psi}(\mathbf{x})}} = \boldsymbol{\psi}(\mathbf{x}) - \mathbb{E}_{\mathcal{P}}[\boldsymbol{\psi}(\mathbf{x})]$. By Jensen's inequality $\|\mathbb{E}_{\mathcal{P}}[\boldsymbol{\psi}(\mathbf{x})]\|_p \leq \mathbb{E}_{\mathcal{P}}[\|\boldsymbol{\psi}(\mathbf{x})\|_p] \leq \alpha$. It follows that $\|\partial f / \partial \mathbf{w}\|_p \leq \|\boldsymbol{\psi}(\mathbf{x})\|_p + \|\mathbb{E}_{\mathcal{P}}[\boldsymbol{\psi}(\mathbf{x})]\|_p \leq \|\boldsymbol{\psi}(\mathbf{x})\|_p + \alpha$. By Definitions 4 and 5, we prove our claim. $\qquad\square$

The requirement that $\mathbb{E}_{\mathcal{P}}[\|\boldsymbol{\psi}(\mathbf{x})\|_p] \leq \alpha$ is also useful in deriving a close-form expresion of the Kullback-Leibler divergence bound.

**Lemma 21.** *Given two continuous factor graphs as in eq.(15), i.e. $\mathcal{P}_1 = p(\cdot|\boldsymbol{\Theta}_1)$ and $\mathcal{P}_2 = p(\cdot|\boldsymbol{\Theta}_2)$, the Kullback-Leibler divergence from $\mathcal{P}_1$ to $\mathcal{P}_2$ is bounded as follows:*

$$\mathcal{KL}(\mathcal{P}_1||\mathcal{P}_2) \leq 2\alpha \|\boldsymbol{\Theta}_1 - \boldsymbol{\Theta}_2\|_p \qquad (16)$$

*Proof.* By invoking Theorem 7, the Lipschitz constant $\overline{K} = \mathbb{E}_{\mathcal{P}_1}[K(\mathbf{x})]$. By invoking Lemma 20, $K(\mathbf{x}) = \|\boldsymbol{\psi}(\mathbf{x})\|_p + \alpha$ and $\mathbb{E}_{\mathcal{P}_1}[\|\boldsymbol{\psi}(\mathbf{x})\|_p] \leq \alpha$. Finally, $\mathbb{E}_{\mathcal{P}_1}[K(\mathbf{x})] = \mathbb{E}_{\mathcal{P}_1}[\|\boldsymbol{\psi}(\mathbf{x})\|_p] + \alpha \leq 2\alpha$. $\qquad\square$

### 4.6 Gaussian Graphical Models

A *Gaussian graphical model* (Lauritzen, 1996) is a Markov random field in which all random variables are continuous and jointly Gaussian. This model corresponds to the multivariate normal distribution.

We first analyze parametrization by using precision matrices. This parametrization is natural since it corresponds to factors graphs as in eq.(15) and therefore conditional independence corresponds to zeros in the precision matrix. We assume that the precision matrix $\boldsymbol{\Omega}$ has bounded norm, i.e. $\alpha \mathbf{I} \preceq \boldsymbol{\Omega} \preceq \beta \mathbf{I}$ or equivalently $\|\boldsymbol{\Omega}^{-1}\|_2 \leq \frac{1}{\alpha}$ and $\|\boldsymbol{\Omega}\|_2 \leq \beta$. This condition holds for Tikhonov regularization as well as for sparseness promoting ($\ell_1$) methods (please, see Appendix B).

**Lemma 22.** *Given the precision matrix $\boldsymbol{\Omega} \succ \mathbf{0}$, the Gaussian graphical model parameterized by $\boldsymbol{\Theta} = \boldsymbol{\Omega}$, $\boldsymbol{\Omega} \in \mathbb{R}^{N \times N}$ with probability density function:*

$$p(\mathbf{x}|\boldsymbol{\Theta}) = \frac{(\det \boldsymbol{\Omega})^{1/2}}{(2\pi)^{N/2}} e^{-\frac{1}{2} \mathbf{x}^{\mathrm{T}} \boldsymbol{\Omega} \mathbf{x}} \qquad (17)$$

*is $(\ell_2, \frac{\|\mathbf{x}\|_2^2}{2} + \frac{1}{2\alpha})$-Lipschitz continuous.*

*Proof.* Let $f(\boldsymbol{\Omega}) = \log p(\mathbf{x}|\boldsymbol{\Theta}) = \frac{1}{2}(\log \det \boldsymbol{\Omega} - N \log(2\pi) - \mathbf{x}^{\mathrm{T}} \boldsymbol{\Omega} \mathbf{x})$. By deriving $\partial f / \partial \boldsymbol{\Omega} = \frac{1}{2}(\boldsymbol{\Omega}^{-1} - \mathbf{x}\mathbf{x}^{\mathrm{T}})$. Therefore $\|\partial f / \partial \boldsymbol{\Omega}\|_2 \leq \frac{1}{2}(\|\boldsymbol{\Omega}^{-1}\|_2 + \|\mathbf{x}\mathbf{x}^{\mathrm{T}}\|_2) = \frac{1}{2}(\|\boldsymbol{\Omega}^{-1}\|_2 + \|\mathbf{x}\|_2^2) \leq \frac{1}{2}(\frac{1}{\alpha} + \|\mathbf{x}\|_2^2)$. By Definitions 4 and 5, we prove our claim. $\qquad\square$

If we use Lemma 21, we will obtain a very loose bound of the Kullback-Leibler divergence where the constant $\overline{K} = \frac{2N\beta^{N/2}}{\alpha^{N/2+1}}$ (please, see Appendix C). Therefore, we analyze the specific case of Gaussian graphical models.

**Lemma 23.** *Given two Gaussian graphical models parameterized by their precision matrices as in eq.(17), i.e. $\mathcal{P}_1 = p(\cdot|\boldsymbol{\Omega}_1)$ and $\mathcal{P}_2 = p(\cdot|\boldsymbol{\Omega}_2)$, the Kullback-Leibler divergence from $\mathcal{P}_1$ to $\mathcal{P}_2$:*

$$\mathcal{KL}(\mathcal{P}_1||\mathcal{P}_2) = \frac{1}{2}\left(\log \frac{\det \boldsymbol{\Omega}_1}{\det \boldsymbol{\Omega}_2} + \langle \boldsymbol{\Omega}_1^{-1}, \boldsymbol{\Omega}_2 \rangle - N\right) \qquad (18)$$

*is bounded as follows:*

$$\mathcal{KL}(\mathcal{P}_1||\mathcal{P}_2) \leq \frac{1}{\alpha} \|\boldsymbol{\Omega}_1 - \boldsymbol{\Omega}_2\|_2 \qquad (19)$$

*Proof.* First, we show that $f(\boldsymbol{\Omega}_1, \boldsymbol{\Omega}_2) = \mathcal{KL}(\mathcal{P}_1||\mathcal{P}_2)$ is Lipschitz continuous with respect to $\boldsymbol{\Omega}_2$. By deriving $\partial f / \partial \boldsymbol{\Omega}_2 = \frac{1}{2}(-\boldsymbol{\Omega}_2^{-1} + \boldsymbol{\Omega}_1^{-1})$. Therefore $\|\partial f / \partial \boldsymbol{\Omega}_2\|_2 \leq \frac{1}{2}(\|\boldsymbol{\Omega}_2^{-1}\|_2 + \|\boldsymbol{\Omega}_1^{-1}\|_2) \leq \frac{1}{2}(\frac{1}{\alpha} + \frac{1}{\alpha}) = \frac{1}{\alpha}$.

Second, since $f$ is Lipschitz continuous with respect to its second parameter, we have $(\forall \boldsymbol{\Omega}) \; |f(\boldsymbol{\Omega}, \boldsymbol{\Omega}_2) - f(\boldsymbol{\Omega}, \boldsymbol{\Omega}_1)| \leq \frac{1}{\alpha} \|\boldsymbol{\Omega}_2 - \boldsymbol{\Omega}_1\|_2$. In particular, let $\boldsymbol{\Omega} = \boldsymbol{\Omega}_1$ and since $f(\boldsymbol{\Omega}_1, \boldsymbol{\Omega}_1) = 0$ and $|f(\boldsymbol{\Omega}_1, \boldsymbol{\Omega}_2)| = f(\boldsymbol{\Omega}_1, \boldsymbol{\Omega}_2)$ by properties of the Kullback-Leibler divergence, we prove our claim. $\qquad\square$

We also analyze parametrization by using covariance matrices (please, see Appendix D). We point out to the reader that this parametrization does not correspond to factors graphs as in eq.(15) and therefore conditional independence does not correspond to zeros in the covariance matrix.

### 4.7 Dynamic Models

The following lemma shows that dynamic Bayesian networks are Lipschitz continuous. Note that dynamic Bayesian networks only impose constraints on the topology of directed graphs, and therefore the extension to the dynamic case is trivial.

**Lemma 24.** *Let $\mathbf{x}^{(t)}$ be the value for variable $\mathbf{x}$ at time $t$, and let $\mathbf{x}^{(t,\ldots,t-L)}$ be a shorthand notation that includes the current time step and the previous $L$ time steps, i.e. $\mathbf{x}^{(t)}, \ldots, \mathbf{x}^{(t-L)}$. Let the set of parents for $x_n^{(t)}$ be $\pi_n \subseteq \{1, \ldots, N\} \times \{0, \ldots, L\}$. Given a $(\ell_p, K)$-Lipschitz continuous conditional probability function $p(x_n^{(t)}|\mathbf{x}_{\pi_n}^{(t,\ldots,t-L)}, \boldsymbol{\Theta})$ for each variable $x_n^{(t)}$, the $L$-order Bayesian network $p(\mathbf{x}^{(t)}|\mathbf{x}^{(t-1)}, \ldots, \mathbf{x}^{(t-L)}, \boldsymbol{\Theta}) = \prod_n p(x_n^{(t)}|\mathbf{x}_{\pi_n}^{(t,\ldots,t-L)}, \boldsymbol{\Theta})$ is $(\ell_p, NK)$-Lipschitz continuous.*

*Proof.* Similar to proof of Lemma 12. $\qquad\square$

The following lemma establishes Lipschitz continuity for conditional random fields.

**Lemma 25.** *Given a feature function with $F$ features $\boldsymbol{\psi}(\mathbf{y}, \mathbf{x}) = (\psi_1(\mathbf{y}, \mathbf{x}), ..., \psi_F(\mathbf{y}, \mathbf{x}))^{\mathrm{T}}$, the conditional random field parameterized by $\boldsymbol{\Theta} = \mathbf{w}$, $\mathbf{w} \in \mathbb{R}^F$ with probability distribution:*

$$p(\mathbf{y}|\mathbf{x}, \boldsymbol{\Theta}) = \frac{1}{\mathcal{Z}(\mathbf{x}, \mathbf{w})} e^{\mathbf{w}^{\mathrm{T}} \boldsymbol{\psi}(\mathbf{y}, \mathbf{x})} \qquad (20)$$

*where $\mathcal{Z}(\mathbf{x}, \mathbf{w}) = \int_{\mathbf{y}} e^{\mathbf{w}^{\mathrm{T}} \boldsymbol{\psi}(\mathbf{y}, \mathbf{x})}$ is $(\ell_p, K)$-Lipschitz continuous.*

*Proof.* Similar to proof of Lemma 19 for discrete random variables, or Lemma 20 for continuous random variables. □

## 5 Experimental Results

First, we show the similarities between the Kullback-Leibler divergence, test log-likelihood and Frobenius norm for some probabilistic graphical models: Gaussian graphical models for continuous data and Ising models for discrete data. Note that if we assume that the test data is generated by a ground truth model, the expected value of the test log-likelihood is the expected log-likelihood that we analyzed in Section 3.

Gaussian graphical models were parameterized by their precision matrices as in eq.(17). We consider Ising models without external field. Therefore, in both cases conditional independence corresponds to parameters of value zero. The ground truth model contains $N = 50$ variables for Gaussian graphical models. For Ising models, since computing the log-partition function is NP-hard, we restrict our experiments to $N = 10$ variables. For each of 50 repetitions, we generate edges in the ground truth model with a required density (either 0.2,0.5,0.8), where each edge weight is generated uniformly at random from $[-1; +1]$. For Gaussian graphical models, we ensure positive definiteness by verifying that the minimum eigenvalue is at least 0.1. We then generate training and testing datasets of 50 samples each. Gaussian graphical models were learnt by the graphical lasso method of (Friedman et al., 2007), and Ising models were learnt by the pseudolikelihood method of (Höfling & Tibshirani, 2009). Figure 1 shows that the Kullback-Leibler divergence, negative test log-likelihood and Frobenius norm behave similarly.

Next, we test the usefulness of our theoretical results that enable us to perform classification, dimensionality reduction and clustering from the parameters of graphical models. We use the CMU motion capture database (`http://mocap.cs.cmu.edu/`) for activity recognition and temporal segmentation. In both cases, we only used the Euler angles for the following 8 markers: left and right humerus, radius, femur and tibia. Our variables measure the change in Euler angles, i.e. the difference between the angle at the current time and 0.05 seconds before. Variables were normalized to have standard deviation one.

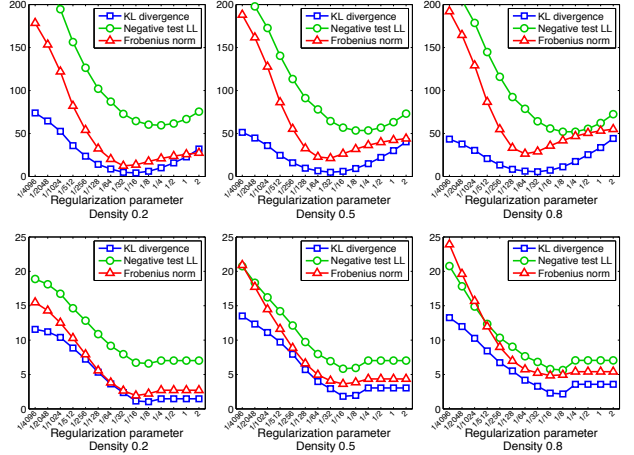

Figure 1: Kullback-Leibler divergence, negative test log-likelihood and Frobenius norm for Gaussian graphical models (top) and Ising models (bottom), for low (left) moderate (center) and high (right) graph density. Note that all the measurements behave similarly.

Table 2: Leave-one-subject-out accuracy for walking vs. running on the CMU motion capture database (chance = 58%).

| Regularization level | 0.001 | 0.01 | 0.1 | 1 |
|---|---|---|---|---|
| $\ell_1$ Covariance | 78 | 78 | 74 | 76 |
| Tikhonov Covariance | 78 | 78 | 78 | 78 |
| $\ell_1$ Precision | 96 | 93 | 90 | 75 |
| Tikhonov Precision | 97 | 96 | 93 | 92 |

For activity recognition, we test whether it is possible to detect if a person is either walking or running from a small window of 0.25 seconds (through the use of classification). The CMU motion capture database contains several sequences per subject. We used the first sequence labeled as "walk" or "run" from all available subjects (excluding 3 pregnant and post-pregnant women). This led to 14 walking subjects and 10 running subjects (total of 21 distinct subjects). From each subject we extracted 3 small windows of 0.25 seconds, at 1/4, 2/4 and 3/4 of the whole sequence. Covariance and precision matrices of Gaussian graphical models were learnt by Tikhonov regularization and the covariance selection method of (Banerjee et al., 2006). Table 2 shows the leave-one-subject-out accuracy for a linear SVM classifier with the parameters of the Gaussian graphical models as features.

For temporal segmentation, we test whether it is possible to separate a complex sequence that includes walking, squats, running, stopping, stretching, jumping, drinking and punching (through dimensionality reduction and clustering). We used the sequence 2 of subject 86 from the CMU motion capture database. We extracted small windows of 0.75 seconds, taken each 0.125 seconds. Each window was labeled as the action being executed in the middle. Precision matrices of Gaussian graphical models were learnt by Tikhonov regularization with regularization level 0.1. We first apply PCA by using the parameters of the Gaussian graphical models as features and then perform k-means clustering with the first 3 eigenvectors. Figure 2 shows the resulting clusters and Table 3 shows the confusion matrix of assigning each window to its cluster.

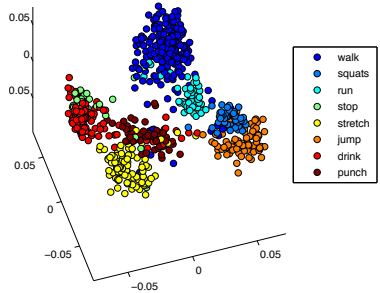

Figure 2: Clusters from a complex sequence of the CMU motion capture database. Each point represents a Gaussian graphical model, the Kullback-Leibler divergence between two points is bounded by the distance between them.

Table 3: Confusion matrix for temporal segmentation from a complex sequence of the CMU motion capture database. Ground truth labels on each row, predicted labels on each column (each row add up to 100%).

|         | walk | squats | run | stop | stretch | jump | drink | punch |
|--------:|------|--------|-----|------|---------|------|-------|-------|
| walk    | 93   |        |     |      | 1       | 2    |       | 4     |
| squats  |      | 87     |     |      |         | 7    |       | 6     |
| run     | 13   |        | 83  |      |         |      |       | 4     |
| stop    | 6    |        |     | 73   | 3       |      | 3     | 15    |
| stretch |      |        |     |      | 70      |      |       | 30    |
| jump    |      | 4      |     |      |         | 96   |       |       |
| drink   |      |        |     | 6    | 4       | 1    | 78    | 11    |
| punch   |      |        |     |      | 16      |      | 4     | 80    |

## 6 Concluding Remarks

There are several ways of extending this research. Lipschitz continuity for the parameterization of other probability distributions (e.g. mixture models) needs to be analyzed. We hope that our preliminary results will motivate work on proving other theoretical properties as well as on learning probabilistic graphical models by using optimization algorithms that rely on Lipschitz continuity of the log-likelihood as the objective function. Finally, while Lipschitz continuity defines an upper bound of the derivative, lower bounds of the derivative will allow for finding a lower bound of the Kullback-Leibler divergence as well as upper bounds for the Bayes error and the expected log-likelihood.

**Acknowledgments.** This work was done while the author was supported in part by NIH Grants 1 R01 DA020949 and 1 R01 EB007530.